\documentclass[sigconf]{acmart}
\AtBeginDocument{%
  \providecommand\BibTeX{{%
    \normalfont B\kern-0.5em{\scshape i\kern-0.25em b}\kern-0.8em\TeX}}}

\usepackage{svg}
\newtoggle{showtodos}
\togglefalse{showtodos}

\usepackage{float}

\renewcommand\footnotetextcopyrightpermission[1]{} 
\setcopyright{none}
\copyrightyear{}
\acmYear{}
\acmDOI{}
\acmConference[Document Intelligence Workshop at KDD]{The Second Document Intelligence Workshop at KDD}{2021}{Virtual Event}
\acmPrice{}
\acmISBN{}

\begin{document}

%

\title[Data-Efficient Information Extraction from Form-Like Documents]{Data-Efficient Information Extraction from \\ Form-Like Documents}

\author{Beliz Gunel}
\authornote{Work done during Google AI research internship, correspondence to bgunel@stanford.edu.}
\affiliation{
 \institution{Stanford University}
 \country{USA}
}
\email{bgunel@stanford.edu}

\author{Navneet Potti}
\affiliation{
 \institution{Google}
 \country{USA}
}
\email{navsan@google.com}

\author{Sandeep Tata}
\affiliation{
 \institution{Google}
 \country{USA}
}
\email{tata@google.com}

\author{James B. Wendt}
\affiliation{
 \institution{Google}
 \country{USA}
}
\email{jwendt@google.com}

\author{Marc Najork}
\affiliation{
 \institution{Google}
 \country{USA}
}
\email{najork@google.com}

\author{Jing Xie}
\affiliation{
 \institution{Google}
 \country{USA}
}
\email{lucyxie@google.com}

\renewcommand{\shortauthors}{Gunel, et al.}

\begin{abstract}
Automating information extraction from form-like documents \textit{at scale} is a pressing need due to its potential impact on automating business workflows across many industries like financial services, insurance, and healthcare. 
The key challenge is that form-like documents in these business workflows can be laid out in virtually infinitely many ways; hence, a good solution to this problem should generalize to documents with \textit{unseen} layouts and languages. A solution to this problem requires a holistic understanding of both the textual segments and the visual cues within a document, which is non-trivial. While the natural language processing and computer vision communities are starting to tackle this problem, there has not been much focus on (1) data-efficiency, and (2) ability to generalize across different document types and languages.

In this paper, we show that when we have only a small number of labeled documents for training ($\sim$50), a straightforward transfer learning approach from a considerably structurally-different larger labeled corpus yields up to a 27 F1 point improvement over simply training on the small corpus in the target domain. We improve on this with a 
\textit{simple multi-domain transfer learning approach}, that is currently in production use, and show that this yields up to a {\em further} 8 F1 point improvement.
We make the case that data efficiency is critical to enable information extraction systems to scale to handle hundreds of different document-types, and learning good representations is critical to accomplishing this.

\end{abstract}


\maketitle

\section{Introduction}
\label{section:introduction}
Given a target set of fields for a particular document type, say, \textit{invoice date} and \textit{total amount} for invoices, along with a small set of manually-labeled documents, the task at hand is to learn to automatically extract these fields from documents with \textit{unseen} layouts and languages. Note that even within the documents of the same \textit{type} and language, say English invoices, same pieces of information may be described in entirely different ways, as each vendor often has their own layout structure. We will refer to this layout structure as \textit{template} for the rest of the paper. Moreover, this information extraction task requires understanding both the textual segments and the visual cues within a document as it aims to generalize to unseen templates across different document types and languages. Hence, the traditional information extraction techniques from webpages, most of which do integrate visual layout information \cite{DBLP:journals/libt/Chowdhury99, DBLP:conf/sigir/CaiYWM04, DBLP:conf/www/YuCWM03, DBLP:conf/kdd/ZhuNWZM06, DBLP:conf/apweb/CaiYWM03}, do not suffice here.

\begin{figure}[H]
  \centering
  \includegraphics[width=0.45\textwidth]{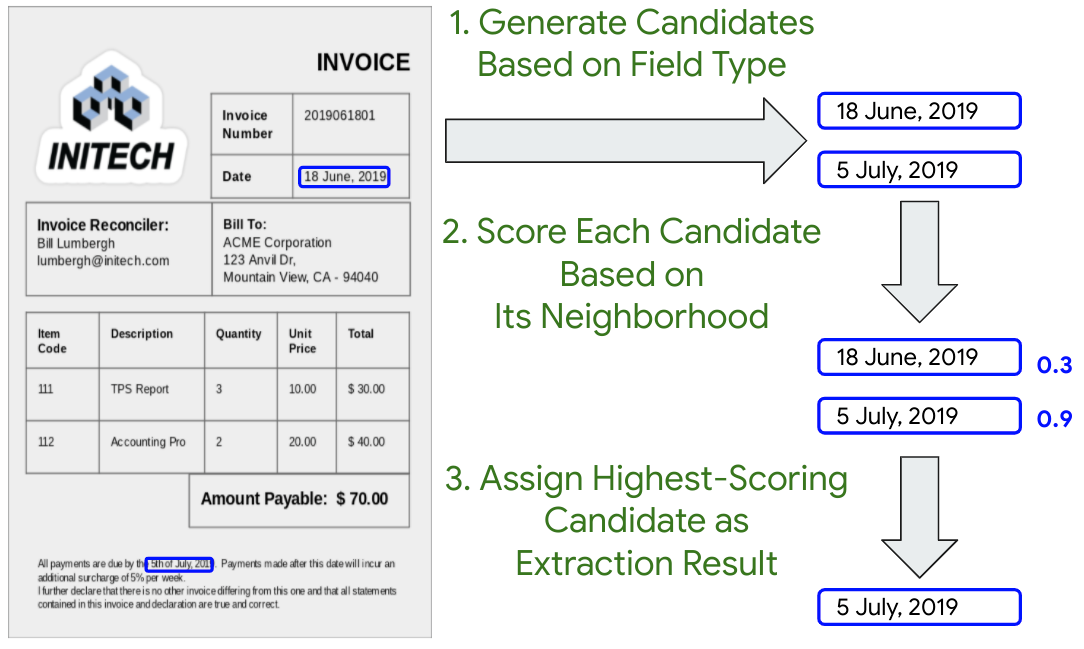}
  \caption{Given a document image and a target schema as inputs, we perform Optical Character Recognition (OCR) on the document image, generate candidates using candidate generators that leverage the existing domain knowledge of working with structured documents, score these generated candidates using a representation learning-based ML model that is described in Section~\ref{sec:glean-overview}, and assign the best candidates to the target fields to produce the final extraction result. }
  \label{fig:glean_pipeline}
  \vspace{-3mm}
\end{figure}


There have been several multi-modal ML-based proposals to tackle this task such as BERTGrid \cite{DBLP:journals/corr/abs-1909-04948}. BERTGrid extends \citet{DBLP:conf/emnlp/KattiRGBBHF18}'s work that previously proposed inputting documents as 2D grids of text tokens to fully convolutional encoder-decoder networks by incorporating pre-trained BERT \cite{DBLP:conf/naacl/DevlinCLT19} text embeddings into the 2D grid representation. Another line of work proposes extending language model pre-training approaches to include the document layout information including \cite{Xu2020LayoutLMPO, Garncarek2020LAMBERTLL}. Although these approaches are promising, training or pre-training time of their pipelines are not only compute- and data-intensive, but also need to be \textit{re-done from scratch} for competitive extraction performance while working with a new considerably structurally different document type or a different language. We would like to point out that in order to fully automate this task, we need to tackle 100s of document types -- and the main cost is data acquisition and labeling for every new language or document type. If we can get to same extraction performance with 10x less data, we effectively cut the cost of developing new extraction models by 10x.

In our work, we borrow the basic architecture from Glean \cite{majumder2020gleanrepresentationlearning, glean2021system}, an information extraction system that uses a factored approach. Glean decouples the task into three stages -- candidate generation, ML-based scoring, and assigning -- as described in Figure~\ref{fig:glean_pipeline}. Glean's design decision of leveraging candidate generators built using standard entity extraction libraries significantly narrows down the search problem for its ML-based scorer model. This way, Glean's ML-based scorer model can fully focus on (1) learning a representation for an extraction candidate that captures its spatial neighborhood on the page, as well as (2) learning the semantics of the target field. We summarize the design of the extraction system and its ML-based scorer model that effectively leverages representation learning \cite{bengio2013representation} ideas in Section~\ref{sec:glean-overview}.

In this paper, we explore whether it is possible to transfer knowledge over to (1) a considerably structurally-different document type in the same language or to (2) same document type in a different language in the context of Glean, when we have several orders of magnitude less labeled documents in the new document type or language. It is, again, crucial to note that this is of great practical importance, as it is often prohibitively expensive to gather and label datasets of large size for every new document type or language.

We build on the hypothesis that form-like documents share a \textit{visual design language} and that the representation learning approach of Glean naturally enables multi-domain training and fine-tuning across different document types and different languages, by its design. In other words, we postulate that a representation learning approach that exploits that \textit{shared visual design language across form-like documents} is precisely why we can effectively transfer knowledge across considerably different domains. The core idea we follow is that we first focus on learning a good encoder for the extraction candidates that understands the spatial relationships and semantics of form-like documents, and then we fine-tune the learned candidate encoder and the field-specific encodings on the new document type or language of interest. 

We show that our \textit{very simple multi-domain transfer learning approach} --that combines extraction candidates from both \textit{source} and \textit{target} domains, and use a common vocabulary across both these domains to train the scorer model followed by fine-tuning the model on the target domain -- enables remarkable data-efficient generalization both from English Invoices to considerably structurally-different new document type Paystubs, and from English Invoices to French Invoices. Our proposed approach consistently improves over both \textit{training from scratch} and simple \textit{transfer learning} baselines up to 1k labeled documents. The value of our proposed approach is particularly impressive in the low data regimes. Specifically, we improve on the training from scratch baseline by up to 35 F1 points, and on the simple transfer learning baseline by up to 8 F1 points for the 50 labeled document case while generalizing to a new document type. Similarly, we improve on the training from scratch baseline by up to 23 F1 points, and on the simple transfer learning baseline by up to 7 F1 points for the 10 labeled document in the target domain case while generalizing to a new language. 


\section{Extraction System Overview}
\label{section:glean_extraction_system}
\label{sec:glean-overview}

\newcommand{\fieldtype}[1]{\textsl{#1}}

We build on Glean \cite{glean2021system} extraction system that is described in  Figure~\ref{fig:glean_pipeline}. Glean takes in a document image and a target schema --that includes target fields and their corresponding types-- as inputs, and performs Optical Character Recognition (OCR). As an example, target schema for \textit{Invoice} document type could include target fields such as \textit{invoice date} of type \fieldtype{date} and total amount of type \fieldtype{price}. Glean supports numerous field types including \fieldtype{integer}, \fieldtype{numeric}, \fieldtype{alphanumeric}, and \fieldtype{currency}, \fieldtype{address}, \fieldtype{phone\_number}, \fieldtype{url}, and other common entity types. Glean leverages an existing library of entity detectors used in Google's Knowledge Graph and are available through a Cloud API~\footnote{https://developers.google.com/knowledge-graph} for all the types described. Open-source entity detection libraries can be used for common types like names, dates, currency amounts, numbers, addresses, URLs, etc.~\footnote{https://cloud.google.com/natural-language/docs/reference/rest/v1/Entity} The candidate generators are designed to be high-recall -- they identify \emph{every} text span in the document that is likely to be of their type. 

Once extraction candidates have been generated, an ML-based Scorer is used to assign a score for each $(\mathit{field}, \mathit{candidate})$ pair that estimates the likelihood that the given extraction candidate is the right extraction value for that field. Multiple fields in the target schema may belong to the same type, say \textit{invoice date} and \textit{due date}, and may therefore share the same set of extraction candidates. An extraction candidate is represented by the text span identified by the candidate generator along with context such as text in its immediate neighborhood to provide the ML-based scorer model with additional relevant features. Finally, candidates for a target field and document are independently scored, and highest scoring candidate for the field is assigned as the final extraction value. Note that, (1) additional business logic specific to a document type can be specified at the assignment stage such as including constraints like \textit{invoice date} must precede \textit{due date} chronologically, (2) precision of the overall extraction system can be adjusted by imposing a minimum score threshold for each field in the target schema.

\begin{figure}[t]
\centering
\includegraphics[width=0.4\textwidth]{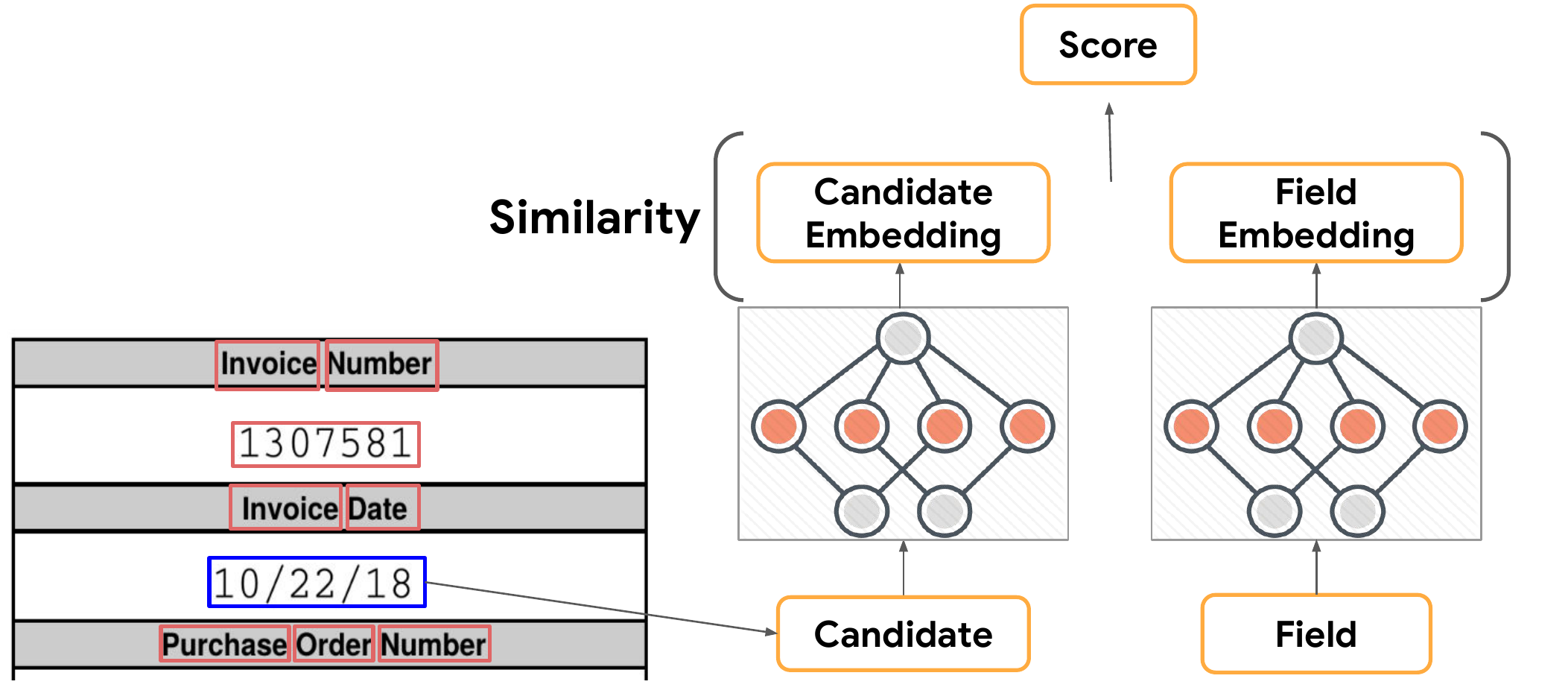}
\centering
\caption{A candidate's score is based on the similarity between its embedding and a field embedding. A date extraction candidate ``10/22/18'' is shown in blue, along with its neighboring tokens, shown in orange.}
\label{fig:scorer}
\end{figure}

A high-level abstraction for the ML-based Scorer is shown in Figure~\ref{fig:scorer}. Modeling specifics of the Scorer architecture is not the focus of this paper, and we refer the reader to \citet{majumder2020gleanrepresentationlearning} for details. In this section, we explain the ML-based Scorer model architecture at a high-level in order for us to qualify \textit{why} the representation learning inspired modeling choices naturally enable multi-domain training and fine-tuning across different document types and different languages. The features of each extraction candidate supplied to the model are its neighboring words and their relative positions, as visualized in Figure~\ref{fig:scorer}. Note that we exclude the candidate's value from the set of features in order to avoid biasing the model towards the distribution of values seen during training, which may not be representative of the entire domain at test time. Model learns a dense representation for each extraction candidate using a simple self-attention based architecture. Separately, in the same embedding space, it learns dense representations for each field in the target schema that capture the semantics of the fields. Based on these learned candidate and field representations, each extraction candidate is scored based on the similarity to its corresponding field embedding. The model is trained as a binary classifier using cross-entropy loss, where the target labels are obtained by comparing the candidate to the ground truth. Please refer to \citet{glean2021system} for how the training data consisting of positive and negative extraction candidates are generated, along with the design decisions to address the data management challenges that arise due to the nature of the problem.

\section{Data Efficiency}
\label{section:data_efficiency}
\subsection{Experimental Setup} \label{sec:expt-setup}

\textbf{Datasets and Evaluation Metrics} Dataset statistics are summarized in Table~\ref{table:datasets}. During data curation, we ensured that no two documents in the corpus share the same template as we aim to learn to extract from \textit{any} template. All of these datasets are proprietary, and hence, are unfortunately not available publicly. Our primary metric is the end-to-end extraction performance measured using Max F1 in the precision-recall curve, which we refer to as the \textit{F1 score}. We always report macro-average F1 score across fields in the target schema. Note that F1 score is affected by the performance of all the parts within the pipeline including the quality of OCR engine and recall of the candidate generators.  All experiments below use a batch size of 256, the Rectified Adam optimizer \cite{liu2019variance} with a learning rate of 0.001. These hyperparameters were found to be optimal using a grid-search. We train the models using an 80-20 train-validation split for up to 25 epochs in each stage, and pick the checkpoint with the best validation ROC AUC for the scorer model, which typically occurs in fewer than 10 epochs.
Experiments on generalization to new language all use a vocabulary consisting of the 2k most frequent tokens occurring in English and French Invoice documents. Experiments on generalization to new document type all use a vocabulary consisting of the 4k most frequent occurring in Paystub and English Invoice documents. All experiments were repeated using 10 different seeded random initializations of the model, and we report the median performance on a fixed, hold-out test set of target domain documents along with error bars that show the variance across these runs.\\

\begin{table}
\centering
\begin{tabular}{lccc}
\hline
& \textbf{$\#$ Docs} & \textbf{$\#$ Fields} & \textbf{Usage} \\
English Invoices & 20k & 12 & Train\\
French Invoices Corpus1 & 5k  & 12 & Train\\
French Invoices Corpus2 & 400 & 12 & Test\\
Paystubs Corpus1 & 10k & 19 & Train \\
Paystubs Corpus2 & 1.5k & 19 & Test\\
\hline
\end{tabular}
\caption{Dataset statistics. $\#$ Docs refer to the number of documents in the corpora, and $\#$ Fields refer to the number of fields in the schema for the given document type.}
\label{table:datasets}
\vspace{-8mm}
\end{table}


\subsection{Multi-Domain Transfer Learning}

\begin{figure*}
\centering
\includegraphics[width=\textwidth]{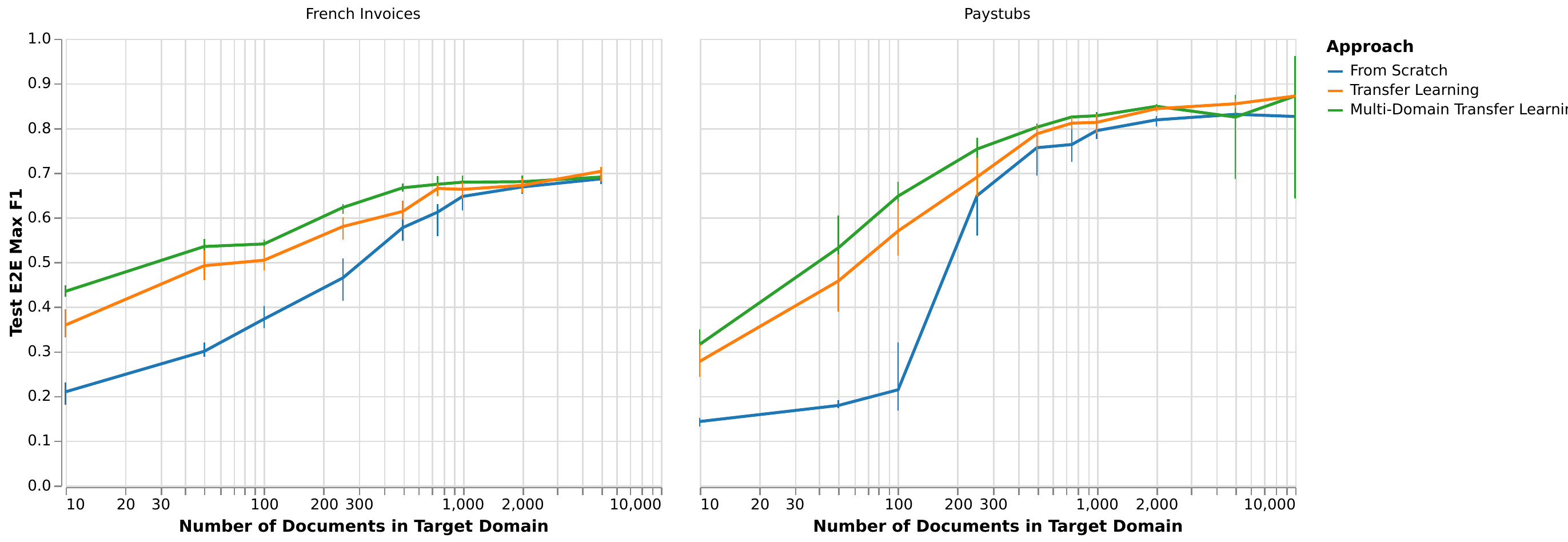}
\centering
\caption{Both learning curves use the backbone architecture described in Section~\ref{sec:glean-overview} and compare \textit{train from scratch} and \textit{transfer learning} baselines to our proposed \textit{multi-domain transfer learning} approach. Left figure shows generalization ability from English Invoices to French Invoices, and right figure demonstrates generalization ability from English Invoices to Paystubs. All fields included in this analysis have above 80$\%$ candidate generation coverage and have at least 40 ground truth label in their corresponding test set. French Invoices document type have 12 and Paystubs have 19 fields in their target schemas. Both figures show median performance at different number of labeled documents for the \textit{target domain} (new document type or language we are trying to generalize to) along with error bars that show variance across 10 different seeds. Proposed \textit{multi-domain transfer learning} approach consistently improves on both baselines up to 1k labeled documents in the \textit{target domain}, and the improvement is particularly significant in the low data regime. \textit{Source domain} for both figures is English Invoices.}
\label{fig:learning_curve}
\end{figure*}

Our proposed \textit{multi-domain transfer learning} approach is a natural extension of the representation learning ideas used in Glean. For the remainder of this paper, we will refer to the \textit{document type} or \textit{language} that we already have enough labeled examples for as \textit{source domain}; and the \textit{document type} or language that we would like to generalize to but not have enough labeled examples as \textit{target domain}. Note that this is of great practical importance, as it is often simply not feasible to gather and label datasets of large size for every new document type or language.

\begin{figure}
\centering
\includegraphics[width=0.5\textwidth]{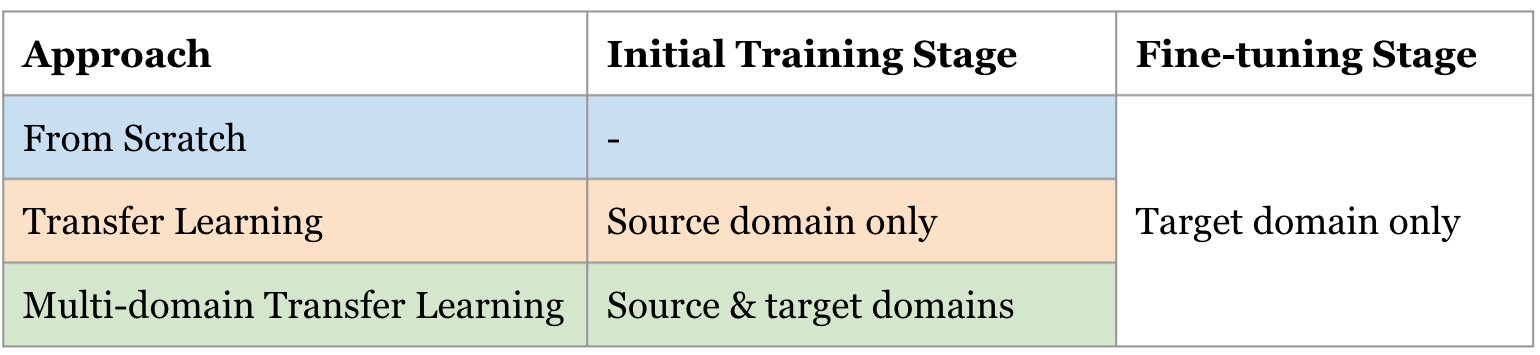}
\centering
\caption{All compared methods use the same backbone described in Section~\ref{sec:glean-overview}. Please refer to Figure~\ref{fig:scorer} for high-level descriptions of Field Embedding and Candidate Encoder, and to \cite{majumder2020gleanrepresentationlearning} for specifics of the ML-based Scorer architecture.}
\label{fig:approaches}
\vspace{-5mm}
\end{figure}

We build on the hypothesis of Glean that form-like documents share a visual design language and that the candidate encoder within the ML-based Scorer we use can effectively learn to represent the domain-agnostic spatial relationships -- between the candidate and its neighbors -- that are critical to understand so that we can have a holistic understanding of form-like documents. Another key observation we make is that the candidate encoder learns embeddings for the neighboring tokens, and these word embeddings, while agnostic to the field, are likely to be specific to the domain used for training. Thus, at Stage 1, we (1) combine the extraction candidates from both the \textit{source} and \textit{target} domains, and (2) use a common vocabulary across both these domains while training Glean's ML-based Scorer model. Significant performance benefit we observe over simple \textit{transfer learning} approaches stems from these two key decisions based on our domain-specific observations, as they make the learned candidate representations \textit{more general}. Field-specific information continues to be encoded in the field embeddings within the same latent space. At Stage 2, we simply fine-tune both the candidate encodings and field embeddings on the \textit{target} domain. Note that this framework can easily be extended to an arbitrary number of \textit{source} and \textit{target} domains. 

Our results shown in Figure~\ref{fig:learning_curve} indicate that our proposed \textit{multi-domain transfer learning} approach enables remarkable data-efficient generalization both from English Invoices to considerably structurally-different new document type Paystubs, and from English Invoices to French Invoices -- consistently improving over both \textit{training from scratch} and simple \textit{transfer learning} baselines up to 1k labeled documents. Summary of all compared methods is described in Figure~\ref{fig:approaches}. Note that \textit{domain} refers to document type in the first described setting, and it refers to language in the second described setting. \textit{training from scratch} baseline simply trains a model using only the labeled examples in the \textit{target} domain. The simple \textit{transfer learning} baseline first trains the ML-based scorer model using the examples in the \textit{source} domain and then fine-tunes the model in the \textit{target} domain. Both learning curves shown in Figure~\ref{fig:learning_curve} use the same backbone described in Section~\ref{sec:glean-overview} and compare \textit{train from scratch} and \textit{transfer learning} baselines with our proposed \textit{multi-domain transfer learning} approach. Left figure shows generalization ability from English Invoices to French Invoices, and right figure demonstrates generalization ability from English Invoices to Paystubs. All fields included in this analysis have above 80$\%$ candidate generation coverage and have at least 40 ground truth label in their corresponding test set. French Invoices document type have 12 and Paystubs have 19 fields in their target schemas. Both figures show median performance at different number of labeled documents for the \textit{target domain} along with error bars. Our proposed \textit{multi-domain transfer learning} approach consistently improves on both baselines up to 1k labeled documents in the \textit{target domain}, and the improvement is particularly significant in the low data regime. \textit{Source domain} for both cases is English Invoices.

The value of our proposed approach is particularly impressive in the low data regimes. Specifically, we improve on the training from scratch baseline by up to 35 F1 points, and on the simple transfer learning baseline by up to 8 F1 points for the 50 labeled document case while generalizing to a new document type. Similarly, we improve on the training from scratch baseline by up to 23 F1 points, and on the simple transfer learning baseline by up to 7 F1 points for the 10 labeled document case while generalizing to a new language. We also would like to point out that (1) source model training takes approximately 45 minutes, converging after 15-25 epochs, and fine-tuning on the target domain approximately takes couple minutes, converging after 1-2 epochs on a single GPU, in contrast to the pre-training based approaches such as BERTGrid model training that takes approximately 1090 minutes converging after 20 epochs based on our own implementation, (2) our proposed multi-domain transfer learning approach is currently in production use.

\section{Related Work}
\label{section:related_work}
\citet{DBLP:conf/emnlp/KattiRGBBHF18} propose inputting documents as 2D grids of text tokens to fully convolutional encoder-decoder networks. \citet{DBLP:journals/corr/abs-1909-04948} incorporate pretrained BERT text embeddings into that 2D grid representation. \citet{Xu2020LayoutLMPO} propose integrating 2D position embeddings and image embeddings, produced with a Faster R-CNN \cite{Ren2015FasterRT} model, into the backbone structure of a BERT language model \cite{DBLP:conf/naacl/DevlinCLT19} and using a masked visual-language loss during pre-training. Similarly, \citet{Garncarek2020LAMBERTLL} propose integrating the 2D layout information into the backbone structure of both BERT and RoBERTa \cite{Liu2019RoBERTaAR}, where they construct layout embeddings using a graph neural network using a heuristically constructed document graph. In contrast to these pre-training based approaches, Glean extraction system that we build on (1) requires several orders of magnitude less labeled training data (2) an order of magnitude less training and inference time for the parts of the extraction system that use ML, without sacrificing the generalization ability to new document types and languages.

\section{Discussion}
\label{section:discussion}
We argued that data-efficiency will be immensely critical as the information extraction systems in production will increasingly need to perform well across \textit{more} document types, \textit{more} languages, and potentially on private customer data -- ideally without sacrificing the generalization ability and the training and inference time of the parts of the extraction system that use ML. We hope that our preliminary results will help start the discussion on the \textit{importance of data-efficiency} while building the next generation information extraction systems tailored to form-like documents for production use. We believe that next big step will be to decrease the labeled document need from $\sim$1k to $\sim$100 for each new (n+1)th document type or language we would like to generalize to.

\bibliographystyle{ACM-Reference-Format}
\bibliography{kdd2021}

\end{document}